\documentclass[letterpaper]{article} 
\usepackage{aaai23}  
\usepackage{times}  
\usepackage{helvet}  
\usepackage{courier}  
\usepackage[hyphens]{url}  
\usepackage{graphicx} 
\usepackage{natbib}  
\usepackage{caption} 
\usepackage{algorithm}
\usepackage{algorithmic}

\usepackage{newfloat}
\usepackage{listings}
\DeclareCaptionStyle{ruled}{labelfont=normalfont,labelsep=colon,strut=off} 
\floatstyle{ruled}
\newfloat{listing}{tb}{lst}{}
\floatname{listing}{Listing}
\usepackage{color}

\title{Weakly Supervised 3D Multi-person Pose Estimation for Large-scale Scenes \\based on Monocular Camera and Single LiDAR}
\author{
   Peishan Cong\textsuperscript{\rm 1,$\dagger$},
   Yiteng Xu\textsuperscript{\rm 1,$\dagger$},
   Yiming Ren\textsuperscript{\rm 1},
   Juze Zhang\textsuperscript{\rm 1},\\
   Lan Xu\textsuperscript{\rm 1,2},
   Jingya Wang\textsuperscript{\rm 1,2},
   Jingyi Yu\textsuperscript{\rm 1,2},
   Yuexin Ma\textsuperscript{\rm 1,2,\footnote{Corresponding author, $\dagger$ Equal contribution}}
}

\affiliations{
   $^{1}$ShanghaiTech University $^{2}$Shanghai Engineering Research Center of Intelligent Vision and Imaging
   \{congpsh,xuyt1,mayuexin\}@shanghaitech.edu.cn
}

\usepackage{bibentry}

\begin{document}

\maketitle

\begin{abstract}

Depth estimation is usually ill-posed and ambiguous for monocular camera-based 3D multi-person pose estimation. Since LiDAR can capture accurate depth information in long-range scenes, it can benefit both the global localization of individuals and the 3D pose estimation by providing rich geometry features. Motivated by this, we propose a monocular camera and single LiDAR-based method for 3D multi-person pose estimation in large-scale scenes, which is easy to deploy and insensitive to light. Specifically, we design an effective fusion strategy to take advantage of multi-modal input data, including images and point cloud, and make full use of temporal information to guide the network to learn natural and coherent human motions. Without relying on any 3D pose annotations, our method exploits the inherent geometry constraints of point cloud for self-supervision and utilizes 2D keypoints on images for weak supervision. Extensive experiments on public datasets and our newly collected dataset demonstrate the superiority and generalization capability of our proposed method. Project homepage is at
\url{https://github.com/4DVLab/FusionPose.git}.
\end{abstract}

\section{Introduction}

3D multi-person pose estimation (3D-MPE) in the wild, especially in large-scale outdoor scenes, has become an increasingly popular research field. It is an essential technique for human motion understanding, which can benefit many downstream real-world applications, including action recognition, sports analysis, surveillance, augmented/virtual reality(AR/VR), autonomous driving, assistive robots, etc. The goal is to localize semantic keypoints of human bodies in 3D space, namely the world coordinate system.

\begin{figure}[t]
    \centering
    \includegraphics[width=1\columnwidth]{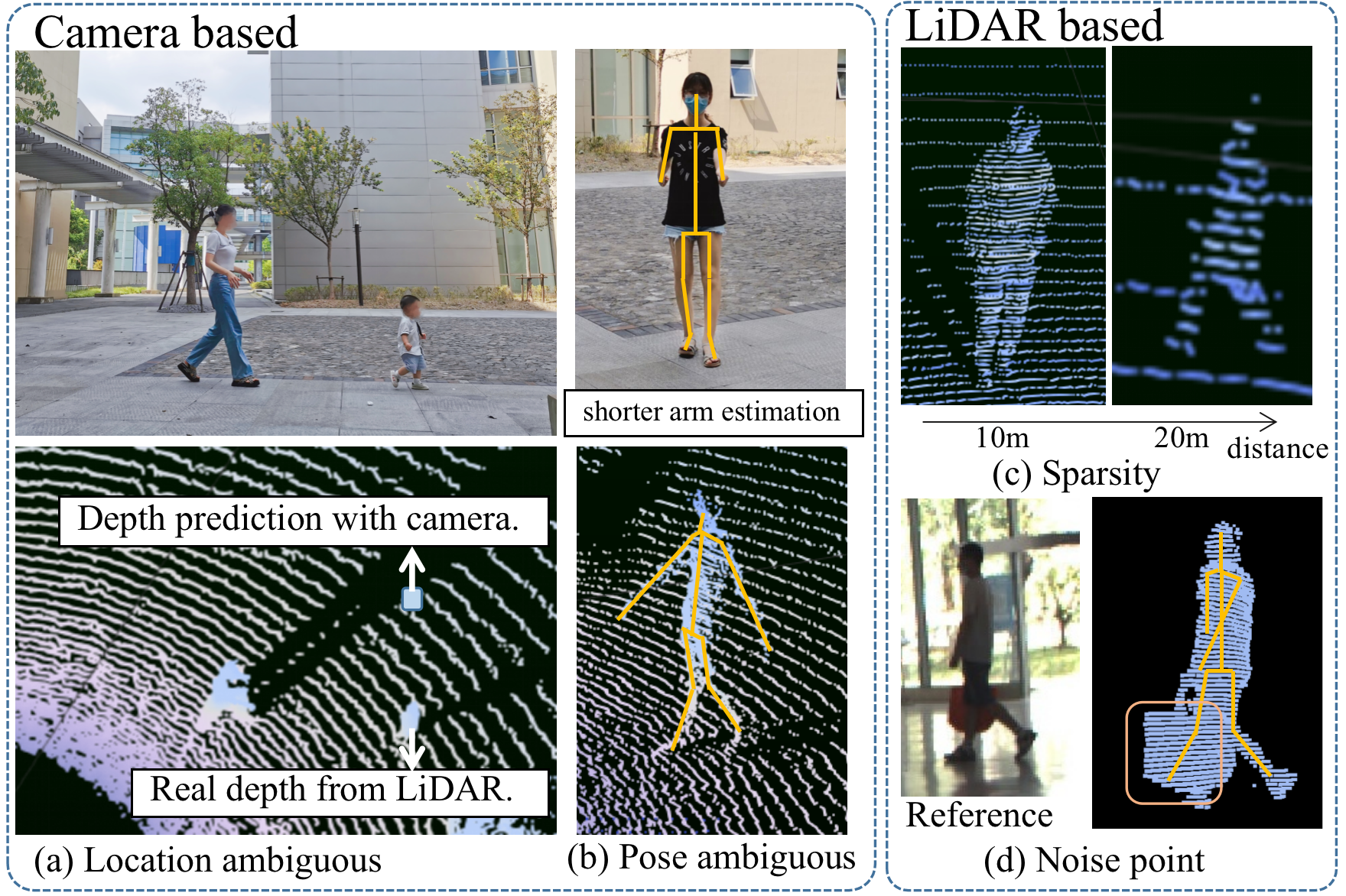}
    \caption{The limitation of 3D-MPE based on monocular camera or single LiDAR. (a) shows the ambiguity of global instance localization caused by the usage of statistics of human body in monocular camera-based methods, which has weak generalization capability for diverse persons. (b) shows the ambiguity of local poses due to the perspective view of image. (c) and (d) illustrate the limitations based on single LiDAR due to the sparse points in the distance and the noise point brought by carry-on objects. FusionPose takes advantage of both sensors to overcome above limitations.}
    \label{fig:camera-lidar}
    \vspace{-3ex}
\end{figure}

Most of previous works~\cite{veges2019absolute,wang2020hmor} solve 3D-MPE based on the monocular camera, which is lightweight and convenient to be set up in general scenarios. However, the problem of depth estimation from monocular camera is ill-posed in essence~\cite{mallot1991inverse}, causing many ambiguous predictions in global localization and local pose estimation, as Figure.~\ref{fig:camera-lidar} shows. Although researchers have proposed plenty of approaches to alleviate the problem, such as using geometric constraints by the prior knowledge of the height or bone length of human body~\cite{zhang2022mutual}, hybrid inverse kinematic constraints~\cite{sun2021monocular,li2021hybrik}, and motion consistency constraints existing in videos~\cite{temporal-zhang2022mixste}. These methods still perform limited due to the mathematically impossible mapping from the perspective view to 3D space. Although the settings of multi-view cameras~\cite{dong2019fast,zhang2021direct} and RGB-D cameras~\cite{zimmermann20183d} are proposed to escape from the trouble, they are not applicable for large-scale scenes due to the deployment difficulties or the physical limitations of sensors (RGB-D camera is available in about 5 meters and usually fail in outdoor scenes.). 

\begin{figure}[t]
    \centering
    \includegraphics[width=1\columnwidth]{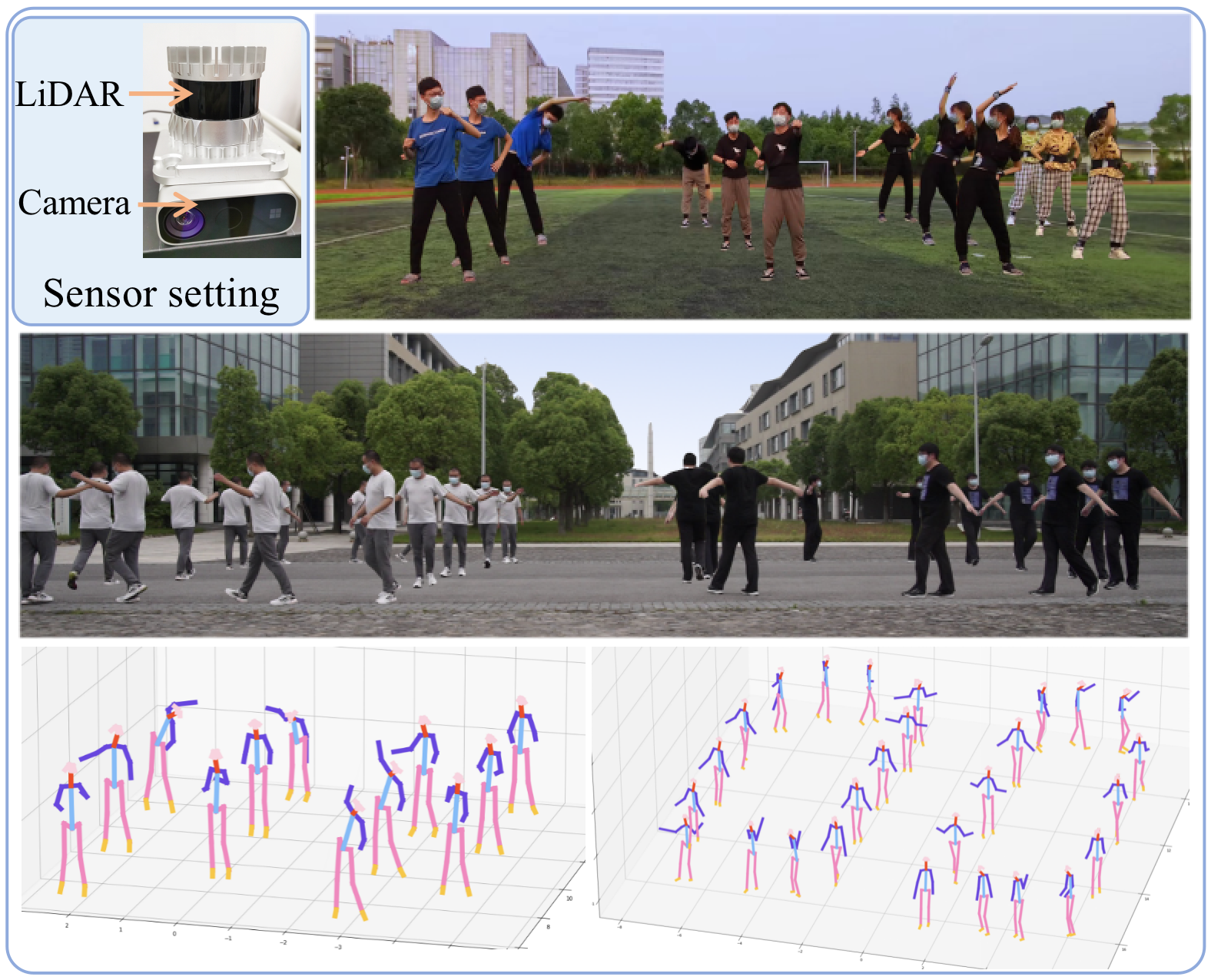}
    \caption{Our sensor setting and 3D multi-person pose estimation results for large-scale scenes. Based on synchronized images and LiDAR point clouds, we can capture continuous global locations and local poses accurately for each person.}
    \label{fig:teasor}
       \vspace{-3ex}
\end{figure}

LiDAR can provide accurate depth information and has been widely-used on autonomous vehicles and robots to perceive large-scale scenes. The effective range for capturing human with recognizable shape and scale could reach about 35m by common 128-beam mechanical LiDAR, making it feasible for 3D-MPE in long-range indoor or outdoor scenes. More importantly, unlike the sensitivity of camera to light, LiDAR could work day and night, which is applicable for most scenarios. Recently, some researchers begin to exploit the LiDAR sensor in human motion capture~\cite{Li2022LiDARCapLM,Zhao2022LiDARaidIP} and has achieved impressive performance. However, the point cloud captured by LiDAR is sparsity-varying along the distance with noise, leading to unstable pose predictions (Figure.~\ref{fig:camera-lidar}). Considering the rich appearance features in images and geometry features in point clouds, ~\cite{furst2021hperl,zheng2022multi} utilize both camera and LiDAR to estimate 3D poses of pedestrians to enhance scene understanding for autonomous vehicles. However, they fuse multi-modal features in a straight forward way and do not employ the temporal information and geometric supervision, leading to unsatisfactory performance on challenging poses, like doing sports. 

We focus on solving 3D-MPE in large-scale scenes based on multi-modal sensors, including monocular camera and single LiDAR. There are two main challenges to overcome. First, the image captured by camera is dense and regular representation and contains the texture feature in the perspective view, while point cloud captured by LiDAR is sparse and unordered representation and provides the depth feature in 3D space. How to taking advantage of multi-modal data from totally different sensors and fuse them in an effective and interpretable way for accurate pose regression is one critical problem. Second, current deep learning-based 3D-MPE methods rely on huge annotated data, which is usually obtained by wearable IMUs devices or multi-view cameras. However, they are not applicable in large-scale scenes due to the drift problem in long distance of IMUs and difficult deployment of multi-view cameras. Manual annotation is expensive and time-consuming. How to conduct 3D-MPE in large-scale scenes without 3D annotations is the other core problem.

In this paper, we propose \textbf{FusionPose}, a novel 3D-MPE approach for large-scale scenes based on the single-LiDAR-camera setting, which has solved above problems. To fully utilize the global semantic feature in images and local geometric feature in point clouds, we present an effective Image-to-Point Attention Fusion (IPAFusion) method to fuse 2D and 3D information. Cross-attention is designed between two modalities to make the network learn the physical correspondence automatically, which can alleviate the dependence on accurate calibration of two sensors and make the fusion process effective and interpretable. To overcome the rely on 3D annotations, we take the best advantage of the self-supervision of the data, including the dynamic motion constraints and high-dimension feature consistency existing in consecutive frames of data, and the geometric constraints of human body points. We also use 2D keypoints generated by mature 2D pose estimation methods to further supervise the estimated 3D keypoints by back projection to image. To facilitate the 3D-MPE research on the multi-modal setting, we collected a new dataset, LiCamPose, in the wild. Extensive experiments show that our method achieves state-of-the-art performance on LiCamPose and other related open datasets. Main contributions of this paper are as follows:
\begin{enumerate}
\item Taking advantage of both LiDAR and camera sensors, we propose a novel method for multi-person 3D pose estimation in large-scale scenes with accurate localization. Specifically, our method is independent of 3D pose annotations.

\item We propose an IPAFusion method to fuse the information of 2D perspective-view images and 3D point cloud, which fully considers global semantic feature and local geometric feature of multimodal data and is free for calibration errors. 

\item We exploit the motion cues and sequential consistency existing in temporal information to enhance the 3D pose estimation.


\item FusionPose achieves state-of-the-art performance on 3D pose datasets, including HybirdCap, 3DPW, STCrowd, and our new collected dataset, LiCamPose. We will release our novel data when the paper is published.

\end{enumerate}
\section{Related Work}

\subsection{Camera-based 3D Human Pose Estimation}
Extensive methods have been proposed for 3D-MPE based on monocular camera. Early works focus
on human-centric tasks without localizing individuals in the actual 3D space. \cite{pavlakos2017coarse} directly regresses the joint positions from input images and \cite{tome2017lifting,martinez2017simple,rogez2019lcr} feed the 2D keypoints into a 2D-to-3D lifting network to estimate 3D poses. To facilitate more real-world applications, researchers pay more attention to the camera-centric 3D-MPE recently. They~\cite{moon2019camera,veges2019absolute,wang2020hmor} usually decouple the problem into the root-relative 2.5D pose estimation and root depth estimation. However, the accurate depth estimation no matter for local keypoints or for objects is the core challenge for 3D-MPE. To address it, some works~\cite{mehta2018single,mehta2020xnect,zhen2020smap,zhang2022mutual} make use of geometry constraints by adding prior knowledge of the human body, such as the height or bone length, in the depth reasoning. 
Based on handcraft assumptions, such methods eliminate many poor results of 3D-MPE but become limited for the scenes with diverse people. Some other methods take advantage of hybrid inverse kinematics of motions~\cite{sun2021monocular,li2021hybrik,sun2022putting,yu2021skeleton2mesh} by using SMPL~\cite{loper2015smpl} parametric human model or explore spatial and temporal relationships by enforcing temporal consistency across consecutive frames~\cite{temporal-arnab2019exploiting,temporal-cheng20203d,temporal-zheng20213d,temporal-zhang2022mixste}. However, the ambiguous depth estimation still exist for the monocular camera setting. Although the multi-camera~\cite{dong2019fast,zhang2021direct,rhodin2018unsupervised,chen2019unsupervised,kocabas2019self,wandt2021canonpose} and RGB-D~\cite{mehta2017vnect,zimmermann20183d,ying2021rgb} settings can, to some extend, alleviate the problem, they are not applicable for the large-scale outdoor scenes.

\subsection{LiDAR-involved 3D Human Pose Estimation}
LiDARs become more and more popular in 3d scene understanding~\cite{Cong_2022_CVPR,Zhu2020SSNSS,zhu2021cylindrical,Yin2020Centerbased3O,han2022licamgait} due to its accurate measurement for the depth information in large-scale scenes, which has boosted the progress of autonomous driving and robotics. Recently, researchers begin to explore the potential usage of LiDAR in fine-grained human motion capture~\cite{Li2022LiDARCapLM,Zhao2022LiDARaidIP} and has made impressive achievements especially for the long-range scenarios. However, LiDAR point cloud has sparse and unordered representation without much texture feature, which usually leads to unstable pose estimations with noise points caused by carry-on objects or clothes. To enhance the perception and understanding for pedestrians in traffic scenarios, \cite{furst2021hperl,zheng2022multi} propose to use both camera and LiDAR to predict the 3D poses of pedestrians. However, they only rely on the 2D keypoint supervision or coarse 3D pseudo labels without considering temporal features, resulting in unsatisfactory results for more complicated actions. Our method leverages the comprehensive feature from images and point clouds, and motion cues in sequences to achieve more robust and accurate pose estimations in more general scenes.


\subsection{Sensor-fusion Approaches for LiDAR and Camera}
There are already many researches about LiDAR-camera-based sensor fusion methods for autonomous driving, which can be classified into three main categories. The first one is point-level fusion strategy~\cite{vora2020pointpainting,wang2021pointaugmenting,zheng2022multi}, which attaches the semantic feature extracted from the corresponding area of image to point, followed by a point cloud-based feature extractor. However, these hard-association methods reply heavily on the sensor calibration and will lose global context information of images. The second one is feature-level fusion strategy~\cite{piergiovanni20214d,MV3D,liang2018deep,ku2018joint} by directing concatenating features from two modalities, which considers the fusion of global context but lacks local geometric corresponding. The third one~\cite{bai2022transfusion,li2022deepfusion,prakash2021multi,liu2022bevfusion} utilizes transformer strategy by constructing queries in BEV space, which dynamically capture the correlations between image and LiDAR features. Such methods work well in detection and segmentation tasks by fusing features in BEV while ignoring fine-grained 3D postures, making them inapplicable for 3D pose estimation tasks. The only two related LiDAR-camera-based 3D-MPE methods~\cite{furst2021hperl,zheng2022multi} directly adopt above point-level and feature-level fusion strategies without specific design for 3D-MPE. In view of the fine-grained feature requirement of 3D-MPE, we propose a soft-association method based on the cross-attention mechanism, which fuse the local geometric features of point cloud with the global context feature of images in an effective manner.

\section{Method}

\paragraph{Problem Definition} 
Given the synchronized image $I$ and point cloud $P$ captured by monocular RGB camera and single LiDAR, our task is to predict the 3D poses $\hat{J}_{3D} \in R^{K \times 3}$ for multiple people in the real world, where $K$ denotes the number of keypoints of the 3D pose representation. Because all sensors are fixed during data capture, the LiDAR coordinate system equals to the world coordinate system, and $\hat{J}_{3D}$ can be projected to image through the intrinsic and extrinsic parameters of sensors. 

\begin{figure*}[ht]
    \centering
    \includegraphics[width=2.10\columnwidth]{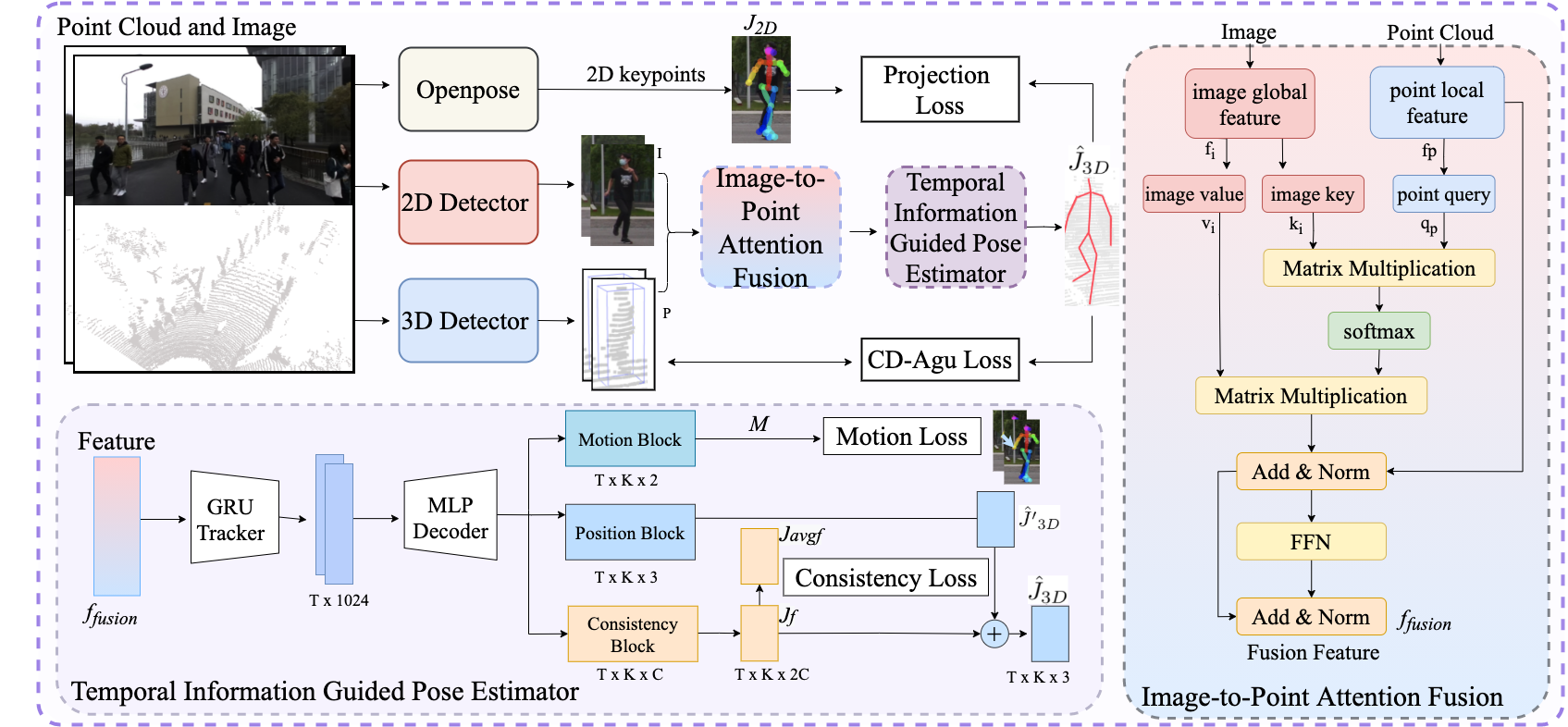}
    \caption{Pipeline of FusionPose. We first obtain the cropped images and point clouds of each person by 2D and 3D detectors. Then the features extracted from multi-modal data are fed in the Image-to-Point Attention Fusion module to get the fused feature with rich texture and geometry information. Temporal Information Guided Pose Estimator is followed to estimate 3D poses by leveraging the temporal features and guidance. Finally, the raw point cloud and 2D keypoints are used for the supervision through shape constraints.
    }
    \label{fig:overview}
       \vspace{-2ex}
\end{figure*}

\paragraph{Overview} Our method is a top-down 3D-MPE method by first detecting persons and then estimating the 3D pose for each person according to the cropped image and point cloud. The whole pipeline of our method is illustrated in Figure \ref{fig:overview}, which contains two important components, including the Image-to-Point Attention Fusion (IPAFusion) module and Temporal Information Guided Pose Estimator. The former fuses the information of two distinct modalities of data to fully use the 3D geometry features of point cloud and the appearance features of images. The latter leverages temporal guidance existing in consecutive data to improve the pose accuracy by learning the dynamic rules of human motions and the pose consistency in high-dimension feature space. Furthermore, we utilize the raw point cloud to supervise the shape and scale of $\hat{J}_{3D}$ and 2D keypoint to weakly supervise the pose by back projection. In the following, we will introduce more details for above modules and losses.


%

\subsection{Pre-processing}
For 2D keypoints $J_{2D} \in R^{K \times 2}$ on images used for supervision, we generate from openpose\cite{cao2017realtime}. For the first-stage detection, we utilize the sate-of-the-art LiDAR-based 3D detector~\cite{Cong_2022_CVPR} and image-based 2D detector (YOLO v5) to process the input and obtain the paired persons in two modalities by projection and matching according to calibration matrix. Tracking is conducted after detection by Hungarian algorithm to find out the person instance correspondence in consecutive frames. Then we crop the images and point clouds by 2D and 3D bounding boxes for the following 3D pose estimation.

\subsection{Image-to-Point Attention Fusion}
Previous fusion methods for LiDAR point clouds and images are designed for detection and segmentation tasks and are applied for autonomous driving. Different from them, 3D pose estimation requires us to pay attention to the fine-grained semantic and geometry features of human bodies. Thus, we propose an effective fusion method for 3D-MPE task, which can automatically learn the corresponding features between images and point cloud to eliminate the sensitivity to sensor calibrations and fully take advantage of global and local information of two modalities.


\paragraph{Point Cloud Feature Extraction}
The low dimensional point cloud input after downsampling $P \in R^{N\times 3}$ ($N=256$ denotes the number of the points) are fed into the PointNet~\cite{qi2017pointnet} encoder to obtain the high-dimensional feature $p=PointNetEncoder(P), p \in R^{N\times 256}$, then we use one layer of self-attention to integrate the global context feature to each point feature:
$$f_p = LN(p+SelfAttention(p)),$$ where $LN$ is layer normalization.

\paragraph{Image Feature Extraction}
We use the pretrained model of HrNet~\cite{wang2020hrnet}  to extract image features, which maintains multi-level resolutions of features and fine-grained local semantic features. The image input $I$ is encoded into high dimensional feature of size $(256, H/8 , W/8)$, where $H$ and $W$ represent the size of input. We flatten the spatial feature and integrate the channel information through Multi-Layer Perception (MLP) to get high-level semantic features: $i= MLP(Flatten({HrNet}(I)))$. One layer of self-attention is also applied to involve the information from global context.
$$f_i = LN(i + SelfAttention(i)).$$



\subsubsection{Cross-attention Fusion}
 Fusing two modal features by direct projection relies heavily on the accurate calibrations of sensors and constrains the correspondence by totally physical mapping. Considering that different parts of the texture feature of images are not equally important to each point of the human body, we design IPAFusion to learn the correspondences between images and point cloud automatically by network, which can fuse features more reasonably and is calibration-free. The \textbf{\emph{point query}} $q_p$ are extracted from high-dimension point feature $f_p$, the \textbf{\emph{image value}} $v_i$ and the \textbf{\emph{image key}} $k_i$ are extracted from global semantic image feature $f_i$. For each query, it conducts a dot product with the image key to get the attention matrix and obtain the correlation from multi-model features. The higher value after the dot product indicates that the point cloud is highly correlated with the corresponding part of the image feature. After softmax normalization, the attention affinity matrix will be multiplied by image value to obtain new point cloud features weighted by the image information.
The weighted point cloud features are then connected with the original point query and pass through two linear layers to obtain $f_{attention}$. The final fusion features $f_{fusion}$ is acquired through FFN in Transformer~\cite{vaswani2017attention}.
$$f_{attention} =  LN(f_p + CrossAttention(q_p,k_i,v_i)), $$
$$f_{fusion} = LN(f_{attention} + FFN(f_{attention})).$$

By this way, IPAFusion can not only automatically learn the correspondences to fuse features of two modalities, but also fully use the global semantic information and local fine-grained geometric feature to boost accurate pose estimation.


\subsection{Temporal Information Guided Pose Estimator}
Human motions are changing continuously with each part of the body moving under specific dynamic constraints. Our method can learn the motion cues in sequential input data by the \textbf{Motion Block} to guide the estimation of more reasonable continuous poses, especially for the occlusion situations, where it is difficult to predict the pose only based on the current frame of data but can be inferred by adjacent poses. Meanwhile, the feature expression of the same keypoint in high dimensional semantic space should be similar, e.g. in high-level feature space, hands even in different frames should keep close and the body center should be far apart from the limbs. We consider the feature consistency in consecutive frames in the \textbf{Consistency Block}.

Figure.~\ref{fig:overview} shows the detailed operations of the temporal information guided pose estimator. First, the fusion features $f^t_{fusion},t\in T$ of $T$ consecutive frames are fed into bi-GRU tracker to extract temporal features. Then, the MLP decoders are followed to predict three different properties of $K$ keypoints, including the motion map $\hat{M}^t \in R^{K \times 2} $ in the motion block, 3D positions in LiDAR coordinate system $\hat{J'}_{3D}^t \in R^{K \times 3}$ in the position block, and high-dimension features $\hat{J_f}^t \in R^{K \times C}$ in the consistency block, respectively.


The motion map ${M}^t$ is calculated the by the difference between each keypoint position at the previous frame and current frame on the image pixel coordinates: $M^t = J_{2D}^{t} - J_{2D}^{t-1}$, the motion prediction is supervised by:
$$L_{motion}=\frac{1}{K}\sum_{j=1}^{K}\left\|\hat{M}_{j}^t-M_{j}^t\right\|,$$
so that the motion block can use the dynamic constraints to assist in more accurate pose estimation.

The consistency block expands the 3D keypoint positions into higher-level space by two extra layers MLP and calculates the temporal consistency loss $L_{consistency}$ to pull the feature $\hat{J}_f^{t}$ of each keypoint to its average feature $J_{avgf}^{t} = \frac{1}{T}\sum_{t = 1}^T J_f^{t}$ cross multiple frames:
$$L_{consistency} = \frac{1}{K}\sum_{j=1}^{K}\left\|\hat{J}^{t}_{f_j}-J^{t}_{avgf_j}\right\|.$$
And then, the feature $J_f^{t}$ is concatenated with the $\hat{J'}_{3D}^t$ and gets the final keypoints $\hat{J}_{3D}^t$ with function $\mathcal{F}$:
$$\hat{J}_{3D}^t = \mathcal{F}(J_f^{t},\hat{J'}_{3D}^t).$$


\subsection{Weakly Unsupervised Training}
2D pose estimation from images has achieved great progress due to huge labeled training data. With the help of 2D poses ${J_{2D}}$ automatically generated by algorithms, we can supervise $\hat{J}_{3D}^t$ by projecting to images according to the transform matrix $\mathcal{T}$. The projected result is represented as $\hat{J}_{2D}^{t} = \mathcal{T}(\hat{J}_{3D}^{t})$. The projection loss is defined as:
$$L_{proj} = \frac{1}{N}\sum_{i=1}^{N}\left\|\mathcal{T}(\hat{J}_{3D}^{t})-J_{2D}^{t}\right\|.$$


In addition, raw point cloud reflect the real shapes, scales, and postures of human body. An accurate estimated 3D pose ought to fit the captured point cloud well. We adopt the Chamfer Distance (CD)~\cite{fan2017point} to measure the similarity between 3D pose and the point cloud.

$$
    L_{CD} (P^t,\hat{J}_{3D}^t) = \frac{1}{\|P^t\|}\sum_{x \in P^t}\min_{y \in \hat{J}_{3D}^t}\|x-y\|_{2}^2 \\$$
$$
      +\frac{1}{\|\hat{J}_{3D}^t\|}\sum_{y \in \hat{J}^t}\min_{x \in P^t}\| y-x \|_{2}^2, 
$$
where $P^t$ denotes the point cloud, $x$ and $y$ represent the 3D coordinates of points.
Only $K$ keypoints is not comparable for N points numerically and geometrically, we further apply linear interpolation on $\hat{J}_{3D}^t$ and calculate the CD\_agu Loss $L_{CD\_agu} (P^t,\hat{J}_{agu}^t)$ with replacing the $J_{3D}$ with $\hat{J}_{agu}^t$ in equation $L_{CD}$, where $\hat{J}_{agu}^t$ is the augmented keypoints.

Then, our network can be trained by the loss $L$ in self-supervised and weak-supervised manner as below:
$$L = \lambda_1 L_{motion} + \lambda_2 L_{consistency} + \lambda_3 L_{proj} + \lambda_4 L_{CD\_agu},$$ where $\lambda$ are hyper-parameters.


%
\subsection{ Implementation Details}

We implement our network using Pytorch 1.10.1 with CUDA 11.3. The point cloud branch is pretrained with simulated data as~\cite{Zhao2022LiDARaidIP} and the image branch utilizes the pretrained feature from HRNet \cite{wang2020hrnet}. The batch size is 8. K is set as 21 as~\cite{cao2017realtime} and T is set as 4 for continuous input frames. 


\begin{figure*}[t]
    \centering
    \includegraphics[width=2.1\columnwidth]{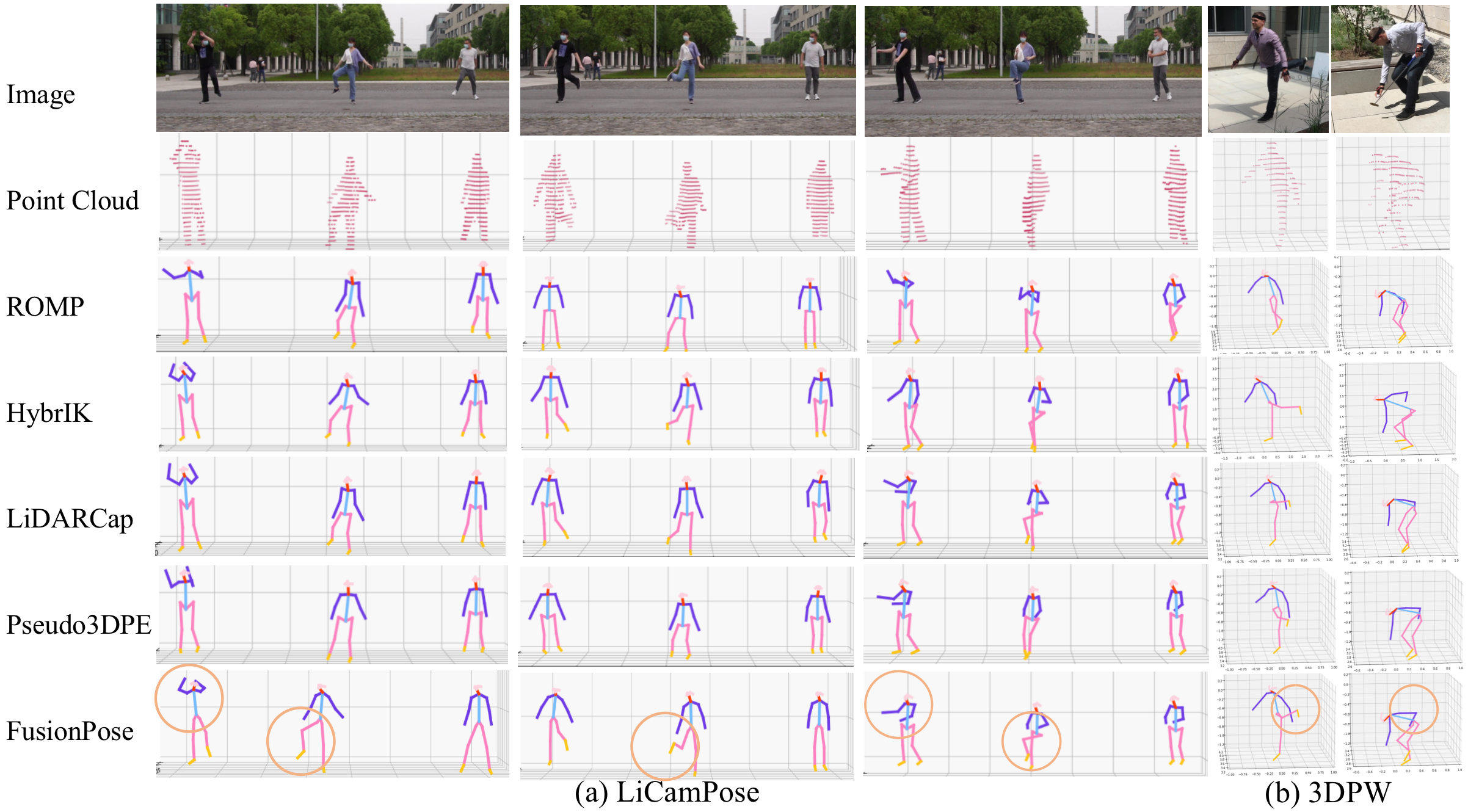}
    \caption{Visualization for local 3D pose results predicted by various methods on HybridCap, LiCamPose and 3DPW. We highlight some parts of estimated 3D poses by circles for detailed comparison. 
    }
    \label{fig:tab 1}
    \vspace{-1ex}
\end{figure*}


\begin{table*}[ht]
  \centering
    \caption{Comparison results on  HybridCap, 3DPW, and LiCamPose. 
    $^{*}$ means fully supervised training mechanism.
    }
    \label{tab:result}
    \setlength{\tabcolsep}{1mm}
\begin{tabular}{ccccccccccccc}

&                              & \multicolumn{3}{c}{HybridCap} & \multicolumn{3}{c}{3DPW}     & \multicolumn{3}{c}{LiCamPose}                          \\\hline
& \multicolumn{1}{c}{Sensor}  &
\multicolumn{1}{c}{PCK$\uparrow$} & \multicolumn{1}{c}{MPJPE$\downarrow$}
 &\multicolumn{1}{c}{CD$\downarrow$} &
 \multicolumn{1}{c}{PCK$\uparrow$} & \multicolumn{1}{c}{MPJPE$\downarrow$} &\multicolumn{1}{c}{CD$\downarrow$}
 &
 \multicolumn{1}{c}{PCK$\uparrow$} & \multicolumn{1}{c}{MPJPE$\downarrow$} &\multicolumn{1}{c}{CD$\downarrow$}
\\\hline
ROMP
& Camera         &     
45.3  & 187.9  &- &53.9&160.6&- &53.2&159.9&-
\\\hline
 HybrIK                      &        Camera                               &                      75.8        &              113.4  & - 
 & 80.2 & 107.1 & -   
 &73.0& 109.2& -
 \\\hline
LidarCap                & LiDAR                     &        
86.5 &    88.1 &  21.2 
&70.7& 119.2 &28.6 
&75.8&119.9&26.8 
\\\hline\hline
Pseudo3DPE   & LiDAR+Camera               &70.7 &130.8 & 28.1  
    
& 73.5 & 116.4 &17.6
& 68.9 & 124.3 & \textbf{19.9}&  
\\\hline
FusionPose         & LiDAR+Camera    
&         \textbf{95.9} & \textbf{75.3} & \textbf{17.4}
& \textbf{83.5} &\textbf{97.7} &\textbf{13.8}
&\textbf{79.7}&\textbf{106.8}&21.7
\\\hline  
{FusionPose*  }       & {LiDAR+Camera }   
&         {95.3} &{75.9} & {17.3}
& {91.3} &{79.2} &{27.5}
&{93.9}&{75.8}&{19.5}
\\\hline                
\end{tabular}
\vspace{-1ex}
\end{table*}

\section{Experiment}
In this section, we first introduce all datasets and evaluation metrics used for experiments and then compare our method with current SOTA methods qualitatively and quantitatively. Extensive ablation studies are conducted for comprehensive assessment of different modules of FusionPose.




\subsection{Dataset}
\noindent \textbf{LiCamPose} is our new collected 3D-MPE dataset in long-range wild scenes with a 128-beam OuSTER-1 LiDAR and a camera, with totally 8,980 frames of synchronized multi-modal data. The ground truth is captured by Noitom Perception Neuron Studio(Noitom PN S). We divide the data half for the training set and half for the testing set. In addition, we collected extra 38,490 frames of data in the same setting but without pose annotations. These data is helpful for unsupervised methods to pretrain their models or validate the performance by visualization. LiCamPose contains various motions, including walking, running, doing KEEP, ball games, dancing, and Taekwondo in different scenes. We protect personal privacy by blurring faces shown in released images.


\noindent \textbf{3DPW}\cite{von20183DPW} and \textbf{Hybridcap}
\cite{liang2022hybridcap} provide images and annotated SMPL models, where 3D poses can be acquired directly, but lack LiDAR point clouds. We follow \cite{cong2021input} to simulate the LiDAR point cloud data in a reasonable manner. Because of the simulation limitation, we select valid data in 3DPW for experiments. 
We follow the official protocol of dataset splitting and will release all processed data to facilitate further evaluation. 



\noindent \textbf{STCrowd}
\cite{Cong_2022_CVPR} is large-scale pedestrian perception dataset with synchronized LiDAR point clouds and camera images, while it doesn't provide the 3D keypoints ground truth. Thus, we only provide the visualization results on this dataset.




\subsection{Evaluation Metric}
 Since LiDAR can provide accurate depth information, we do not compare the depth estimations and only compare local poses with other methods. For evaluation metrics, we use 1) \textbf{PCK$\uparrow$}: percentage of correct keypoints that the normalized distance between the key point and its groundtruth is less than the set threshold (150mm)
position error in millimeters; 2) \textbf{MPJPE$\downarrow$}: mean per root-relative joint position error in millimeter; 3)  \textbf{CD$\downarrow$}: the chamfer
distance between predict keypoints and raw point cloud in millimeter. 
\subsection{Performance Analysis}
We compare with three kinds of SOTA methods for 3D-MPE, including monocular camera-based ROMP~\cite{sun2021monocular} and HybrIK~\cite{li2021hybrik}, LiDAR-based LidarCap~\cite{li2022lidarcap}, and LiDAR-camera multi-sensor-based Pseudo3DPE~\cite{zhang2022mutual}. We run their released code with provided parameters. The results on HybridCap, 3DPW, and LiCamPose are shown in Table \ref{tab:result}.

The camera-based methods are pretrained on MSCOCO, Human3.6M\cite{ionescu2013human3} and MPI-INF-3DHP \cite{mehta2017monocular} in a full-supervision manner and then directly infer on the test data of these datasets. 
For LiDAR-based method, LiDARCap, we pretrain it on LiDARCap dataset with 3D annotations and show results by finetuning using our self-supervision losses. 
Pseudo3DPE and our method are weakly supervised methods with only 2D pose annotations. 

Pseudo3DPE uses pseudo 3D labels for supervision by projecting 3D points into 2D to find adjacent points in the vicinity of 2D key points, the label quality suffers from low-resolution images and accurate calibration. 
Compared with Pseudo3DPE, we get large improvement, illustrating the efficiency and generalization of our feature-fusion method and loss designs. Compared with camera-only and LiDAR-only methods, FusionPose has more accurate pose estimation even without any supervised pretrain stage. 
The visualization results are illustrated in Figure \ref{fig:tab 1}. Since camera-based methods are not good at global localization, we drag the 3D pose to the global position offered by LiDAR for local posture comparison. Due to the depth ambiguous, camera-based methods have poor performance on anterior and posterior amplitude of the limbs. While the LiDAR-based method is affected by the noise of the point cloud and generate rough local postures in some cases. Pseudo3DPE utilizes pseudo 3D labels projected by 2D points, which is not accurate enough and easy affected by calibration errors. Our method have superior performance on both local pose and depth estimation. In particular, we can see that our performance by weakly-supervised training is comparable to that by fully-supervised training, which further demonstrates the effectiveness of FusionPose. More results of our method on the whole large-scale scenes are shown in Figure.~\ref{fig:teasor}.

\begin{table}[]
 \centering
    \caption{Ablation experiment for different fusion methods.}
    \vspace{-1ex}
    \label{tab:ablation1}
    \setlength{\tabcolsep}{0.7mm}
\begin{tabular}{ccccccccc}
                               & \multicolumn{3}{c}{HybridCap}& \multicolumn{3}{c}{3DPW}                             \\\hline     
     & \multicolumn{1}{c}{PCK$\uparrow$} & \multicolumn{1}{c}{MPJPE$\downarrow$} & \multicolumn{1}{c}{CD$\downarrow$}& \multicolumn{1}{c}{PCK$\uparrow$} & \multicolumn{1}{c}{MPJPE$\downarrow$} & \multicolumn{1}{c}{CD$\downarrow$}
     \\\hline     
Point-RGB
&        78.1                      &   112.6  & 21.36  & 73.6 &   110.7 & 15.39                  \\

 PixelFusion&80.3&107.8&20.05
   & 75.6 & 106.9 & 14.91                   \\
LocalFusion  &74.7 & 118.2& 22.85 & 75.3 & 108.7 & 15.96\\ 
GlobalFusion      &        71.1                      &    126.4 & 26.57   & 78.6 & 104.0 & 15.71                   \\
IPAFusion                    
 & \textbf{90.7}   &  \textbf{89.5} & \textbf{19.70}  & \textbf{79.2} & \textbf{103.4} &     \textbf{14.49}   \\\hline    
\end{tabular}
\end{table}

\begin{figure}[]
    \centering
    \includegraphics[width=1.0\columnwidth]{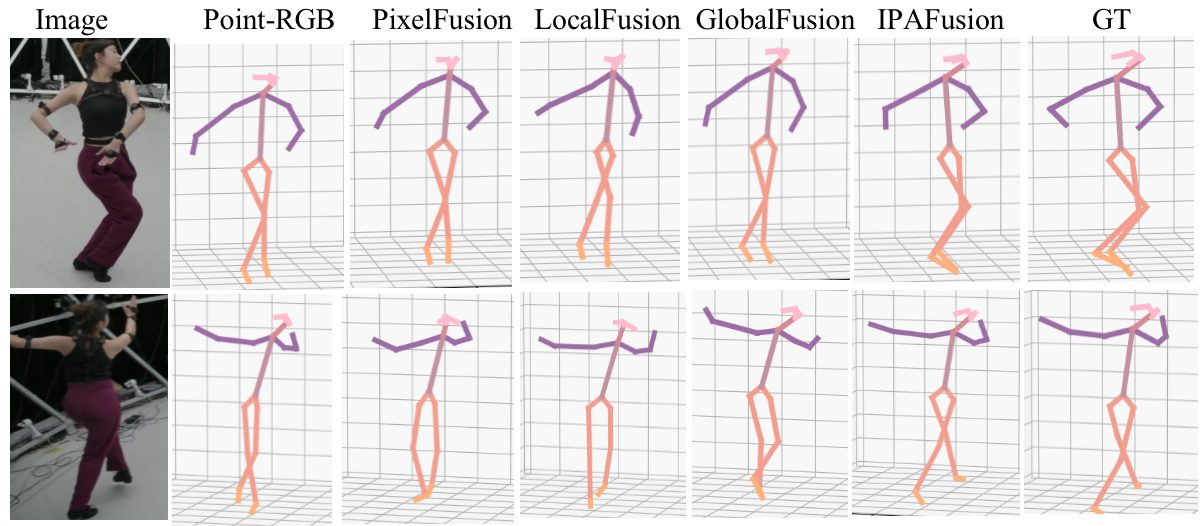}
  
    \caption{Comparison of different fusion methods on HybridCap Dataset. The last column is the ground truth.}
    \label{fig:fusion-exp}
       \vspace{-3ex}
\end{figure}

\subsection{Ablation Study}
In this section, we first validate the effectiveness of our sensor-fusion module by comparing with other fusion approaches. And then we conduct the ablation study for each loss designed in our method.

\noindent \textbf{Ablation Study for IPAFusion:} We demonstrate the superiority of our IPAFusion model by replacing it with other fusion methods. We conduct the comparison on the basic network of FusionPose without extra temporal and CD supervision. Commonly used fusion strategies for images and point clouds in the perception area are as follows:

\noindent\textbf{Point-RGB} appends the raw representation of LiDAR point with corresponding RGB color according to calibration matrix. 
\noindent\textbf{PixelFusion} adds a k-dimensional image feature vector as a supplementary feature for each LiDAR point by projection.
\noindent\textbf{LocalFusion} concatenates k-dimensional image feature vector to the corresponding high dimensional point feature for each point. Above three methods will get a feature-enhanced point cloud and then pass a point cloud-based backbone for further feature extraction. Their performances are sensitive to the sensor calibration, which is not stable in outdoor scenes.
\noindent\textbf{GlobalFusion} directly concatenates the global image features with the global point cloud features. It is a totally high-level fusion strategy. Such method lacks actual mapping from two modal data, which downshifts the learning process and is not feasible for fine-grained pose tasks.
\noindent\textbf{IPAFusion} integrates local high-dimensional geometric features of point cloud with global appearance features of images and automatically learn the projection by the cross-attention mechanism, which is calibration-free and maintain more detailed features. Table \ref{tab:ablation1} and Figure \ref{fig:fusion-exp} show that IPAFusion is significantly better than other fusion methods.

\begin{figure}[]
    \centering
    \includegraphics[width=1\columnwidth]{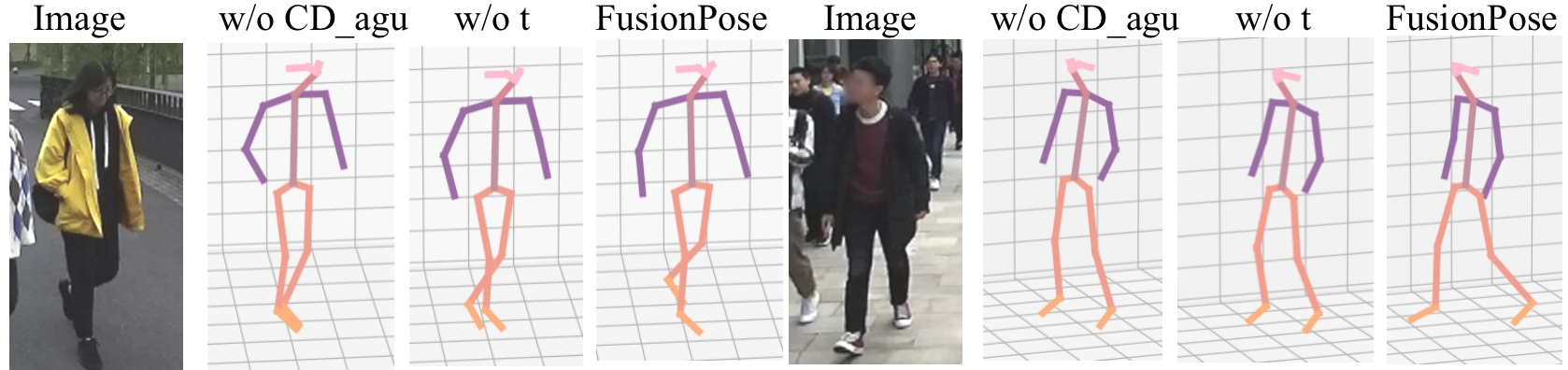}
    \caption{Ablation results of FusionPose components on STCrowd. w/o CD\_agu represents eliminating the CD\_agu loss and w/o t represents eliminating temporal supervision (motion loss and consistency loss).}
    \label{fig:temporal-exp}
    \vspace{-3ex}
\end{figure}

\noindent \textbf{Ablation Study for Loss functions:}
We verify the loss functions of our method quantitatively and qualitatively, as Table.~\ref{tab:ablation2} and Figure.~\ref{fig:temporal-exp} shows. And we also test on single image input, getting PCK 76.1 on LiCamPose (79.7 for full model). With the optimization of CD\_agu loss with geometry constraints and temporal information guided model with dynamic constraints, the performance of FusionPose gets improved.
\subsection{More Analysis}

\noindent \textbf{Analysis for occlusion:} We further conduct experiments by randomly occluding 60\% of LiDAR point cloud of the person, and the PCK performance of our method on LiCamPose, HybridCap, 3DPW dataset are are $78.7, 93.8, 80.0$ (Original results are $79.8,95.9,83.5$) with small drop. However, our top-down method definitely relies on the detection and tracking pre-processing results. For severe external occlusions, such as multi-person hugging, the failure of detection and tracking will directly cause the failure of our method.

\begin{table}[ht]
 \centering
    \caption{Ablation experiment for different components of FusionPose on HybridCap. IPA is the original IPAFusion baseline. CB and MB are consistency block and motion block, respectively,  and CDA means CD\_Agu optimization.}
     \label{tab:ablation2}
    \setlength{\tabcolsep}{0.5mm}
\begin{tabular}{lcccccccccccccccccc}
\hline   
\multicolumn{1}{c}{}        &\multicolumn{1}{c}{PCK$\uparrow$}         &\multicolumn{1}{c}{MPJPE$\downarrow$}         &\multicolumn{1}{c}{CD$\downarrow$}    
         \\\hline
     IPA                 & 90.7                            &           89.5 & 19.7       
                  
\\
        IPA+CB               &      93.6                        &          82.3 & 18.8               
   
\\
IPA+CB+MB                  &    95.4                          &   77.0 & 17.6 
\\IPA+CB+MB+CDA             &       \textbf{95.9}                       &            \textbf{75.3}               & \textbf{17.4} 

\\\hline
\end{tabular}
\end{table}

\begin{table}[ht]
\centering
    \caption{Ablation study on different numbers of points on LiCamPose.}
    \label{tab:ablation-density}
    \begin{tabular}{ccccc}
    \hline
    Points Number &256  & 128  & 64   & 32   \\
    PCK & 79.3 & 76.7 & 73.6 & 71.5\\
    \hline
    \end{tabular}
\end{table}

\noindent \textbf{Analysis for point cloud sparsity:}
We randomly downsample the point cloud in LiCamPose to various densities, the result is shown in Table \ref{tab:ablation-density}. PCK get a little drop with the point number decreases. Meanwhile, we divide our dataset LiCamPose into two subsets according to the distance to LiDAR, and get PCK $76.0$ in range 10-15m and PCK $73.7$ in 15-20m. It illustrates the robustness of our method with sparsity-varying scenes. For the LiDAR sensor we use, the number of points on the person is about 50 when the person is 25m far from the sensor, and we can get appreciable results. For more high-resolution LiDAR sensors, our method can work for longer ranges.

\section{Conclusion}

We propose a new 3D-MPE method for large-scale scenes based on single LiDAR and monocular camera. To fully use the appearance features of images and geometry features of LiDAR point clouds, we propose an effective sensor-fusion method to extract rich and fine-grained local pose features. In particular, our method does not require any 3D annotation by using motion cues and geometry constraints. Extensive experiments show our method achieves state-of-the-art performance on new collected dataset and open datasets.

\section{Acknowledgements}

This work was supported by NSFC (No.62206173, No.61976138), the National Key Research and Development Program (2018YFB2100500), STCSM (2015F0203-000-06), SHMEC (2019-01-07-00-01-E00003), Shanghai Sailing Program (No.22YF1428700, No.21YF1429400), and Shanghai Frontiers Science Center of Human-centered Artificial Intelligence (ShangHAI).

\bibliography{aaai23}

\end{document}